\pdfoutput=1
\documentclass[11pt]{article}
\usepackage{amssymb}  
\usepackage[linesnumbered,ruled,vlined]{algorithm2e}
\usepackage{amsmath}  
\usepackage{xcolor} 
\definecolor{nonFed}{rgb}{1.0, 0.949, 1.0} 
\definecolor{Fed}{rgb}{0.902, 1.0, 0.902} 
\usepackage{booktabs} 
\usepackage{array}    

\usepackage[final]{acl}

\usepackage{times}
\usepackage{latexsym}

\usepackage[T1]{fontenc}

\usepackage[utf8]{inputenc}

\usepackage{microtype}

\usepackage{inconsolata}

\usepackage{graphicx}
\usepackage{epstopdf}
\usepackage{subfigure}

\title{
FLEKE: Federated Locate-then-Edit Knowledge Editing

}

\author{
  Zongkai Zhao$^1$\thanks{Equal Contribution}, Guozeng Xu$^1$\footnotemark[1], Xiuhua Li$^{1}$\thanks{Corresponding Author}, Kaiwen Wei$^2$\footnotemark[2],
  Jiang Zhong$^2$ \\
  $^1$School of Big Data \& Software Engineering, Chongqing University, China \\ $^2$College of Computer Science, Chongqing University, China\\
  {\tt\footnotesize \{zongkaizhao, archipes\}@stu.cqu.edu.cn, \{lixiuhua, weikaiwen, zhongjiang\}@cqu.edu.cn}
}

\begin{document}
\maketitle
\begin{abstract} 
Locate-then-Edit Knowledge Editing (LEKE) is a key technique for updating large language models (LLMs) without full retraining. However, existing methods assume a single-user setting and become inefficient in real-world multi-client scenarios, where decentralized organizations (e.g., hospitals, financial institutions) independently update overlapping knowledge, leading to redundant mediator knowledge vector (MKV) computations and privacy concerns.
To address these challenges, we introduce $\textbf{F}$ederated $\textbf{L}$ocate-then-$\textbf{E}$dit $\textbf{K}$nowledge $\textbf{E}$diting (FLEKE), a novel task that enables multiple clients to collaboratively perform LEKE while preserving privacy and reducing computational overhead. To achieve this, we propose FedEdit, a two-stage framework that optimizes MKV selection and reuse.
In the first stage, clients locally apply LEKE and upload the computed MKVs. In the second stage, rather than relying solely on server-based MKV sharing, FLEKE allows clients retrieve relevant MKVs based on cosine similarity, enabling knowledge re-edit and minimizing redundant computations.
Experimental results on two benchmark datasets demonstrate that FedEdit retains over 96\% of the performance of non-federated LEKE while significantly outperforming a FedAvg-based baseline by approximately twofold. Besides, we find that MEMIT performs more consistently than PMET in the FLEKE task with our FedEdit framework. Our code is available at \href{https://github.com/zongkaiz/FLEKE}{https://github.com/zongkaiz/FLEKE}. 

\end{abstract}

\section{Introduction}

Locate-then-Edit Knowledge Editing (LEKE) has emerged as a key paradigm for updating large language models (LLMs) by directly identifying and modifying model parameters associated with newly acquired knowledge, eliminating the need for costly full-model retraining \cite{r5,r2,r25}. It has proven effective in mitigating hallucinations \cite{r27}, detoxifying outputs \cite{r24}, and improving factual recall \cite{r26}.

\begin{figure}[t]
  \includegraphics[width=\columnwidth]{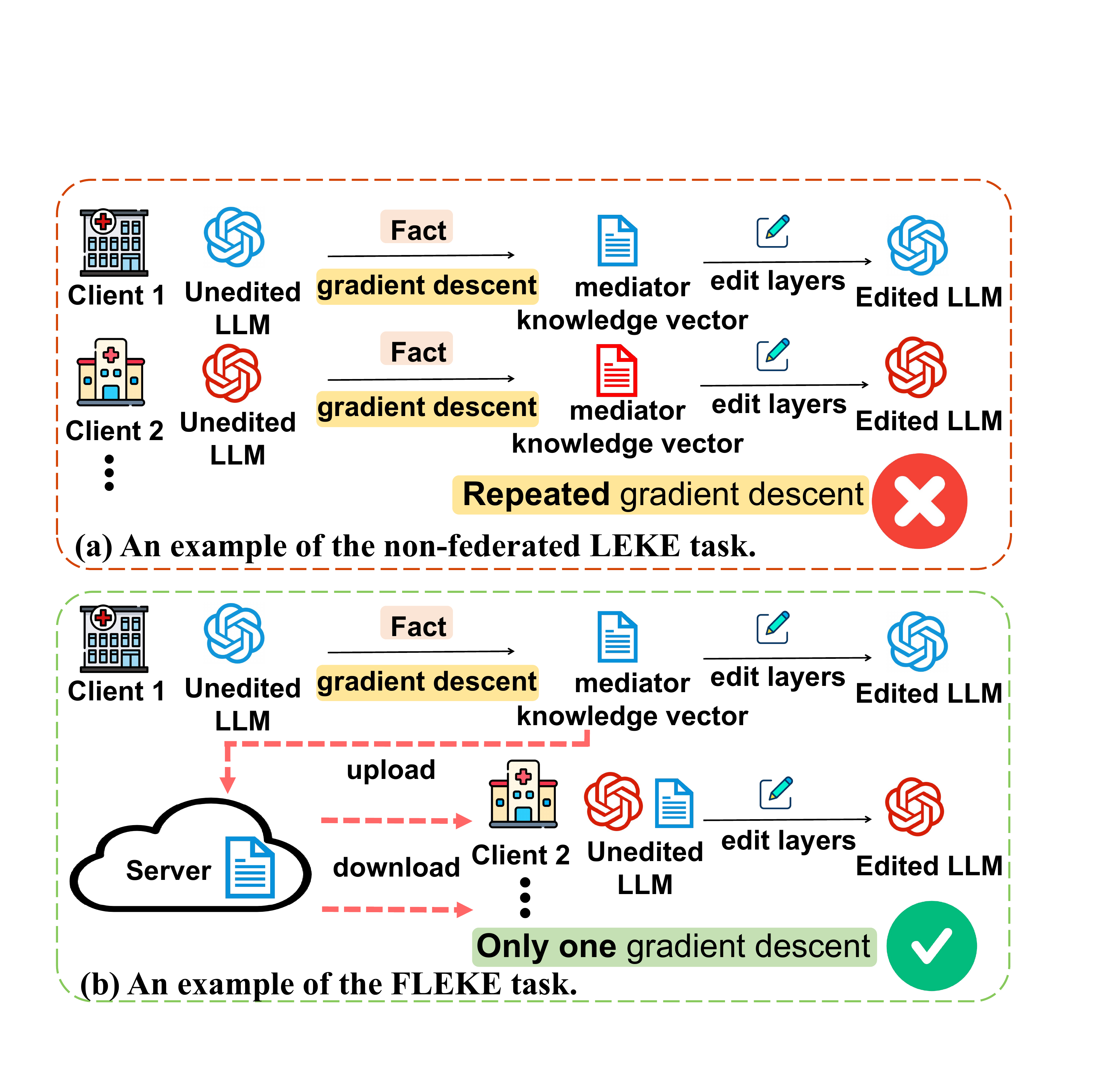}
  \caption{ Comparison between (a) non-federated LEKE and (b) the proposed FLEKE task, where the former requires the computation of the mediator knowledge vector multiple times for the same knowledge through gradient descent, while the latter computes it only once.
  }
  \label{fig:hospital example}
  \vspace{-0.1in}
\end{figure} 

However, existing methods are all conducted in single-client scenarios. Considering real-life applications, as shown in Fig.~\ref{fig:hospital example}(a), traditional LEKE methods suffer from redundant gradient descent computations of mediator knowledge vectors (MKVs) \cite{r5,r2,r7}, leading to inefficiencies in knowledge updates, especially for organizations within the same domain (e.g., different hospitals) that often process overlapping information. This not only exacerbates these inefficiencies but also raises privacy concerns due to data sharing \cite{r22, r23}.
To mitigate these issues, federated learning (FL) \cite{r14,r15,r16} enables collaborative model training while preserving data privacy, making it particularly suitable for such sensitive domains like healthcare and finance.

To extend the LEKE task to federated settings, we propose a new task: \textbf{F}ederated \textbf{L}ocate-then-\textbf{E}dit \textbf{K}nowledge \textbf{E}diting (FLEKE), enabling multiple clients to collaboratively edit knowledge while reducing computational costs and preserving privacy. As shwon in Fig.~\ref{fig:hospital example}(b), In FLEKE, each client runs the LEKE algorithm locally to generate MKVs representing knowledge updates. These MKVs are uploaded to a central server, where they are stored and shared, preventing redundant computations. When a predefined time slot arrives, clients retrieve relevant MKVs from the server to refine their knowledge. 

\begin{figure}[t!]
  \centering
  \includegraphics[width=0.48\textwidth]{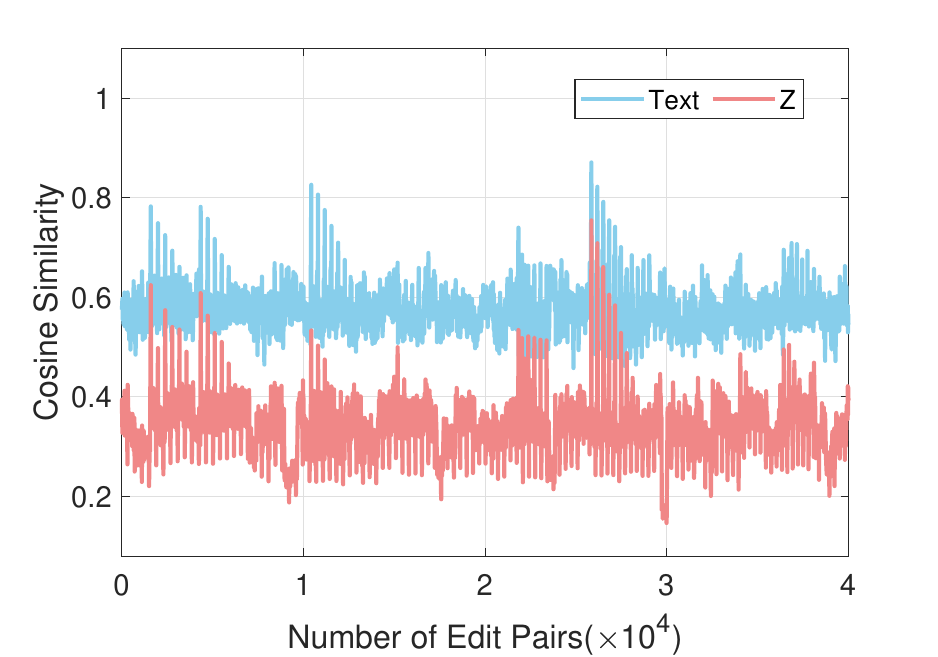}
  \caption{Cosine similarity between core text in zsRE dataset and the corresponding $z_i$ vectors.}
  \label{fig:cos}
  \vspace{-0.15in}
\end{figure}

To accomplish FLEKE, several critical challenges need to be addressed:
(1) \textit{How to define MKVs for client update}: Unlike traditional LEKE in federated settings, which requires frequent recalculation of MKVs for the same knowledge, FLEKE computes them only once and shares them across clients via a central server, so selecting appropriate MKVs for upload is crucial. They must effectively encode essential knowledge while remaining computationally efficient.
(2) \textit{How to retrieve relevant MKVs for client download}: Efficient retrieval is crucial to minimize computational and storage costs while ensuring clients access only the most relevant MKVs. A key issue is dynamically selecting MKVs that best match each client’s needs, balancing retrieval efficiency and knowledge quality.

To address the first challenge, we explored various representations for mediator knowledge vectors (MKVs) and found $z_i$ vectors introduced in \citet{r2} to be suitable. As shown in Fig.~\ref{fig:cos}, our analysis of the zsRE dataset \cite{r4} identified 2,000 edit pairs where the cosine similarity between the core text and corresponding $z_i$ vectors exceeded 0.65, indicating shared information. Statistical validation confirmed a strong positive correlation, with a Pearson coefficient of 0.74 \cite{r28}. 
These findings show that $z_i$ vectors effectively encode original knowledge while improving computational efficiency, making them well-suited as MKVs.

To address the second challenge, we propose \textbf{FedEdit}, it operates in two stages: first, at predefined intervals, clients apply existing LEKE algorithms to update multiple layers of their models, uploading the computed MKVs to the server. Then, in the re-editing stage, clients periodically evaluate the similarity between their local data and the vectors stored on the server. And a re-editing condition is established, if the similarity meets a predefined threshold, the server’s vectors can be reused for further editing, allowing clients to refine their models without redundant computations. 

We reorganize two large-scale counterfactual datasets zsRE and C{\small{OUNTER}}F{\small{ACT}} \cite{r5} to simulate the FLEKE task,.
Extensive experiments on GPT-J (6B) \cite{r20} and GPT-NeoX (20B) \cite{r1} show that even in the FLEKE setting, the proposed FedEdit method retains at least 96\% of the performance of state-of-the-art methods in non-federated environments.
The key contributions of this work are summarized as follows:

1) We introduce FLEKE, a task enabling multi-client collaborative knowledge editing in dynamic scenarios. To the best of our knowledge, this is the first work to apply LEKE in the federated setting.

2) We introduce FedEdit, a two-stage editing framework designed to improve multi-client editing efficiency for related knowledge, where a re-editing condition is established to efficiently select mediator knowledge vectors from the server. 

3) We reorganize the zsRE and C{\small{OUNTER}}F{\small{ACT}} datasets to simulate FLEKE. Experimental results show that, under FLEKE conditions, FedEdit achieves performance at or above 96\% of that of state-of-the-art methods in non-federated settings. 

\section{Related Work}

\textbf{Locate-then-Edit Knowledge Editing.} 
The locate-then-edit approach in knowledge editing identifies and modifies specific weights in pre-trained models to achieve desired outputs \cite{locate-then-edit-1, locate-then-edit-2}. Various methods have been proposed within this framework. ROME \cite{ROME} updates the feedforward network to encode new knowledge, while MEMIT \cite{MEIMT} extends this for large-scale editing. PMET \cite{PMET} enhances MEMIT’s performance with a residual attribution strategy. Additionally, ROME \cite{ROME} and MEMIT \cite{MEIMT} use input prompts to locate and edit knowledge neurons. However, existing works do not address multi-client scenarios and multi-editing tasks. In this paper, we propose a federated locate-then-edit knowledge editing framework to improve editing efficiency in such settings. \\
\textbf{Federated Learning in LLMs.}
Research on combining large language models (LLMs) and federated learning (FL) primarily focuses on pre-training and prompt engineering \cite{fl-llm}. Pre-trained models, trained on large datasets, serve as a foundation for FL, significantly reducing training time \cite{fl-llm-time1, fl-llm-time2} and helping address data and system heterogeneity \cite{fl-llm-non-iid}. Some studies incorporate pre-trained models into FL frameworks for various tasks \cite{pre-fl1, pre-fl2}. Prompt-based techniques have shown strong performance in LLMs \cite{promptfl}. The pFedPT framework personalizes models efficiently using personalized prompts \cite{personlization}, while DiPrompT \cite{DiPrompT} applies adaptive prompts to tackle domain generalization challenges in FL. To the best of our knowledge, this is the first work to apply FL for optimizing LEKE in LLMs.

\section{Method}\label{method: method}
In this section, we provide a detailed introduction to the FLEKE task and the FedEdit framework. First, we discuss the relationship among the hidden states of each Transformer layer in the LLM and the relationship between the hidden states and the input in section~\ref{method: Preliminaries}, which is essential for calculating the MKVs. Next, we introduce the FLEKE task and explain its connection to the LEKE task in section~\ref{method: FLEKE}, and we also analyze how to optimize and solve it. Then, we focus on solving the LEKE task and extracting the relevant knowledge vector in section~\ref{method: LEKE}. Finally, we propose the FedEdit framework to address the FLEKE task in section~\ref{method: FedEdit}.
\subsection{Preliminaries}\label{method: Preliminaries}
This section introduces the foundational concepts of autoregressive and decoder-only LLM models, focusing on the relationship between the hidden states of each Transformer layer and the input. These foundations are essential for calculating the MKVs.

Autoregressive and decoder-only LLMs denoted as $\mathcal{F}_{\theta}$ encode input sequences $x$ into $z$ token sequences $x_{1},...,x_{z}$, which are processed through $L$ Transformer decoder layers. The probability of the next token $x_{z+1}$ is computed as:
\begin{equation}
\small
\begin{aligned}
\mathcal{F}_{\theta}(x_{1},...,x_{z}) 
& =\mathrm{softmax}\left(W_{\mathrm{E}}\gamma\left(h_{z}^{L-1}+a_{z}^{L}+m_{z}^{L}\right)\right) \\
& =\mathbb{P}\left(x_{z+1}|x_{1},...,x_{z}\right),
\end{aligned}
\end{equation}
where $W_{\mathrm{E}}$ and $\gamma$ are the embedding matrix and layer norm, respectively, and $a_{z}^{L}$, $m_{z}^{L}$ are the hidden states of the MHSA and FFN at the $L$-th layer. $a_{j}^{l}$, $m_{j}^{l}$ for the $j$-th token at layer $l$ are:
\begin{equation}\small\begin{aligned}
 & a_{j}^{l}=W_{O^{\mathrm{MHSA}}}^{l}\mathrm{MHSA}^{l}\left(\gamma\left(h_{1}^{l-1},h_{2}^{l-1},...,h_{j}^{l-1}\right)\right), \\
 & m_{j}^{l}=W_{O^{\mathrm{FFN}}}^{l}\sigma\left(W_{I}^{l}\gamma\left(h_{j}^{l-1}\right)\right),
\end{aligned}\end{equation}
where $W_{O^{\mathrm{MHSA}}}$ and $W_{O^{\mathrm{FFN}}}$ are weights for MHSA and FFN, and $\gamma$ is the activation function.
\subsection{FLEKE Task Formulation}\label{method: FLEKE}

In this section, we present the FLEKE task and explain its connection to the traditional LEKE task. The FLEKE refers to the collaborative execution of the LEKE task by multiple clients in a federated scenario. 
Assuming that each client c has a fact data set $\mathcal{E}_c^t$ to be edited in time slot $t$, the goal of FLEKE is to insert the fact data $\mathcal{E}$ of all clients by editing the internal parameters of LLM.
Overall, for each client $c$ between predefined time slots, FLEKE optimizes an objective function to obtain target weights \cite{r2}:
\begin{equation}\small
\begin{aligned}
    W_c^t \triangleq \underset{\tilde{W_c^t}}{\operatorname{argmin}} \Bigg( 
    \sum_{i=1}^n& \left\| \tilde{W_c^t} k_{ic}^t - v_{ic}^t \right\|^2 \notag \\
     +&\sum_{i=n+1}^{n+u} \left\| \tilde{W_c^t} k_{ic}^t - v_{ic}^t \right\|^2 
    \Bigg).
\end{aligned}
\end{equation}
Here, $k_{ic}^t$ $\triangleq$ $k_{ic}^{tl}$ and $v_{ic}^t$ $\triangleq$ $v_{ic}^{tl}$ represent the sets of keys and values, respectively, encoding the subject-related knowledge in the $l$-$th$ layer at time $t$ on client $c$. The term $\sum_{i=1}^n \left\| \tilde{W_c^t}k_{ic}^t - v_{ic}^t \right\|^2$ indicates that we aim to retain $n$ pieces of knowledge, while $\sum_{i=n+1}^{n+u} \left\| \tilde{W_c^t}k_{ic}^t - v_{ic}^t \right\|^2$ suggests that we intend to modify a much larger number of knowledge pieces, denoted as $u$ $\gg$ 1. Here, the keys and values are represented as matrices stacked horizontally: $\begin{bmatrix} k_{1c}^t \mid k_{2c}^t \mid \cdots \mid k_{nc}^t \end{bmatrix} \triangleq K_c^t$ and $\begin{bmatrix} v_{1c}^t \mid v_{2c}^t \mid \cdots \mid v_{nc}^t \end{bmatrix} \triangleq V_c^t$. The target weight $W_c^t$ is the sum of the original weight $\tilde{W_c^t}$ and the incremental weight $\Delta_c^t$, i.e., $W_c^t=\tilde{W_c^t}+\Delta_c^t$. Based on the derivation from MEMIT \cite{r2}, the formal expression for the incremental weight is given as:
\begin{equation}\label{Eq:Delta}\Delta_c^t=R_c^tK_c^{t^T}(C_0+K_c^tK_c^{t^T})^{-1},\end{equation}
where $R_c^t\triangleq V_c^t-\tilde{W_c^t}K_c^t$ represents the residual between the values $V_c^t$ (namely the target knowledge representations) corresponding to the keys $K_c^t$ of the target knowledge and the client $c$ model's original knowledge $\tilde{W_c^t}K_c^t$. $C_0\triangleq\lambda\mathbb{E}_k\left[kk^T\right]$ is an estimate of the set of previously memorized keys obtained through sampling, and $\lambda$ is a hyperparameter that balances the degree of model modification and preservation.

\begin{figure}[t]
  \includegraphics[width=0.485\textwidth]{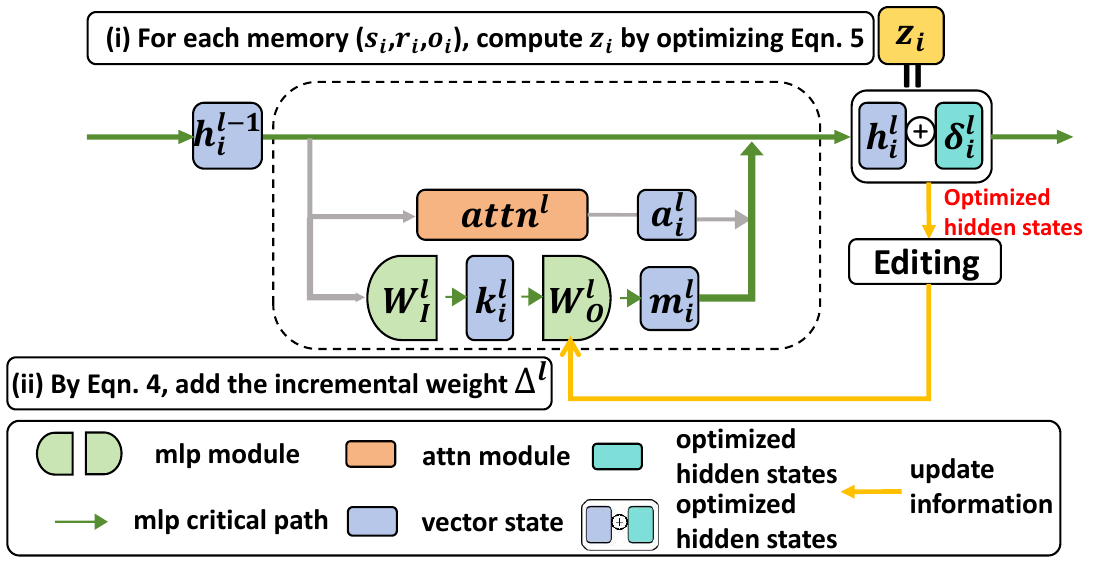}
  \caption{The overview of the classic LEKE method named MEMIT~\cite{r2}.}
  \label{fig:LEKE}
  \vspace{-0.1in}
\end{figure}
\subsection{LEKE}\label{method: LEKE}
This section delves into the LEKE method, emphasizing how knowledge updates are performed across multiple layers of the Transformer. For instance, as shown in Fig.~\ref{fig:LEKE}, MEMIT \cite{r2} employs optimized transformer layer hidden states to perform subtle updates on the FFN weights. In contrast, PMET \cite{r7} simultaneously optimizes the transformer component hidden states of both MHSA and FFN, but only applies the optimized TC hidden states to the FFN. In this paper, we take MEMIT as an example of a LEKE method and further elaborate on its approach to updating multiple layers in the FLEKE task. Specifically, we calculate the target knowledge set of the first and last critical layer $L_0 = \min(\mathcal{R}), L = \max(\mathcal{R}) $. For each edit $(s_{ci}, r_{ci}, o_{ci}) \in \mathcal{E}_c$ (sucbject $s$, relation $r$, object $o$) on client $c$, we (i) compute $z_{ci}$ to replace $h_{ci}^L$ such that adding $\delta_{ci} \triangleq z_{ci} - h_{ci}^L$ to the hidden state at layer $L$. Then, for each layer, we (ii) modify the MLP at layer $l$ by spreading $\Delta_c^{tl}$ over layer $l$.

(i) \textbf{Computing $z_{ci}$.} For the $i$-th edit on client $c$, $z_{ci}$ is derived by optimizing the residual vector $\delta_{ci}$ via gradient descent:
\begin{equation}\small
\begin{aligned}
z_{ci}=&h_{ci}^L+\underset{\delta_{ci}}{\operatorname*{\operatorname*{argmin}}}\frac{1}{P}\sum_{j=1}^P\\&-\log\mathbb{P}_{\mathcal{F}_{c}(h_{ci}^L+=\delta_{ci})}\left[o_{ci}\mid x_{cj}\oplus p(s_{ci},r_{ci})\right].\label{eq:zi}
\end{aligned}
\end{equation}

In words, we optimize $\delta_{ci}$ to maximize the client $c$ model’s prediction accuracy for the desired object $o_{ci}$, given a set of factual prompts $\{x_{cj}\oplus p(s_{ci},r_{ci})\}$ that concatenate random prefixes $x_{cj}$ to a templated prompt to aid generalization across contexts. $\mathcal{F}_{c}(h_{ci}^L+=\delta_{ci})$ indicates that we modify the transformer execution by substituting the modified hidden state $z_{ci}$ for $h_{ci}^L$.

(ii) \textbf{Spreading $\Delta_c^{tl}$ over layer $l$.} We follow the same algorithm steps as MEMIT that are presented in Algorithm \ref{alg:MEMIT} in Appendix \ref{sec:appendixA}. Next, we'll mainly describe how to implement our FedEdit framework with the update step.

\begin{figure}[t]
  \includegraphics[width=0.48\textwidth]{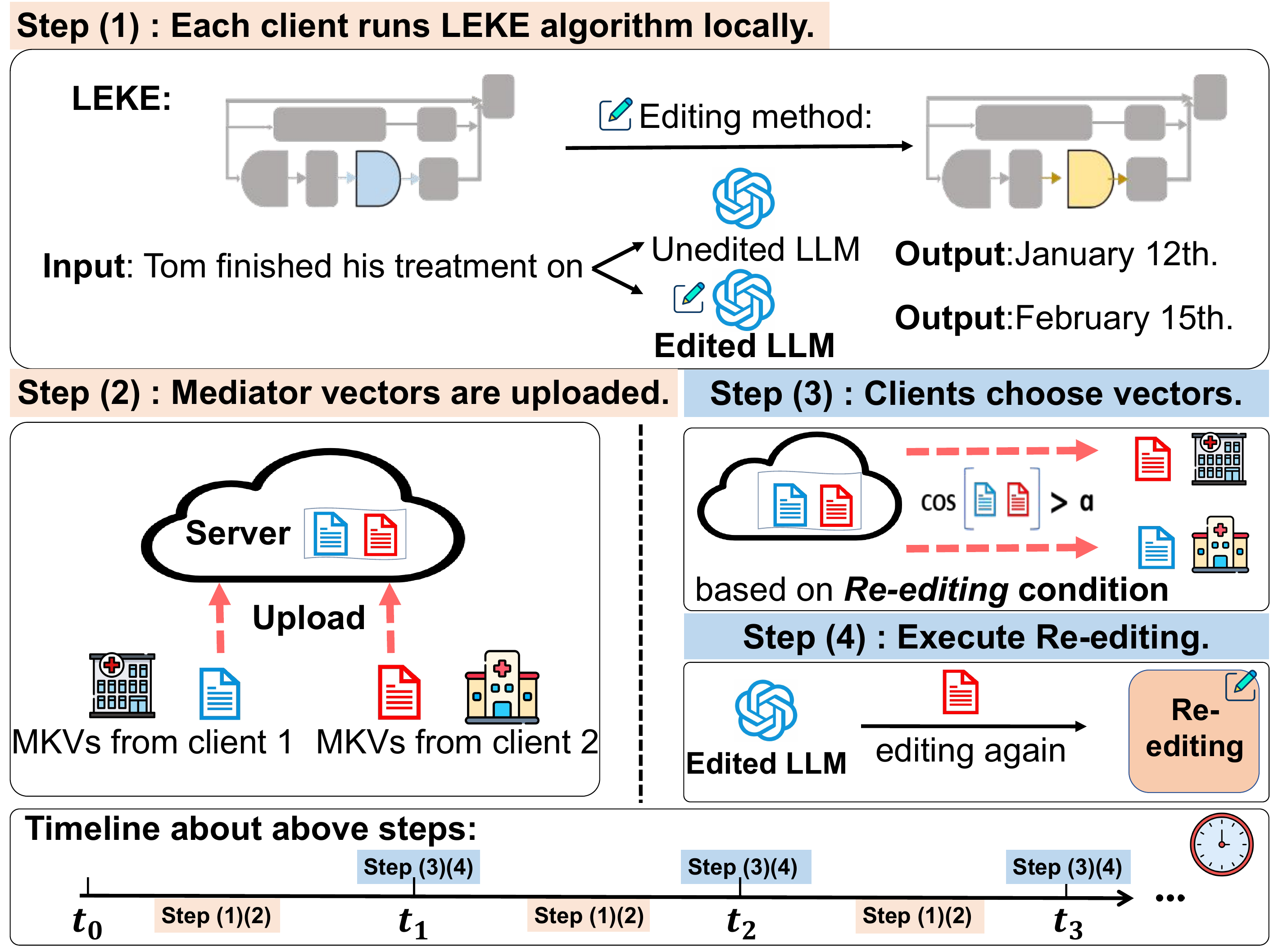}
  \caption{The workflow of the proposed FedEdit.}
  \label{fig:overview}
  \vspace{-0.1in}
\end{figure}
\subsection{FedEdit Framework}\label{method: FedEdit}
In this section, we propose the FedEdit framework to address the FLEKE task in a federated setting. The framework is designed to adapt LEKE tasks to a federated scenario, where each client interacts with the server and collaboratively edits the knowledge. As shown in Fig.~\ref{fig:overview}, the workflow of the FedEdit framework is as the following steps:

Step (1): Starting at $t = 0$, each client runs the Edit algorithm locally, which can be any LEKE method. In this paper, we select MEMIT \cite{r2} and PMET \cite{r7}. This process generates the MKVs.

Step (2): The MKVs are then uploaded to the server.

Step (3): At a predetermined time slot, each client selects some MKVs from the server according to the re-editing conditions defined later.

Step (4): If at least one vector is chosen by the client, it continues editing on the model.

On the timeline, steps (1) and (2) occur within the intervals between given time slots, while steps (3) and (4) are executed when the predetermined time slot is reached.

\begin{algorithm}[t]
\small
\caption{FedEdit} \label{alg:FedEdit}
\KwIn{similarity threshold $\alpha$, the number of time slots $m$, records $\mathcal{E}$, unedited model $\mathcal{M}$}

Initialize $client$ $\gets$ [$client_1$, $client_2$, $\dots$, $client_n$], $server$ $\gets$ [], $t$ $\gets 0$, $T$ $\gets$ [$t_1$, $t_2$, $\dots$, $t_m$], $selected\_z$ $\gets$ [] \;

$t$ begins to increment \;

\For{$c \in client$ \textbf{in parallel}}{
    ${edited\_model}_c$, $Z_c^{t_i} \gets \textbf{Edit}(model_c, \mathcal{E}_c)$ \;
    $server$.append($Z_c^{t_i}$) \;
    \If{$t \in T$ \textbf{and} \textbf{Select\_$Z$}($server$, $Z_c^{t_i}$) $\neq \emptyset$}{
        \textbf{Edit}(${edited\_model}_c$, \textbf{Select\_$Z$}($server$, $Z_c^{t_i}$)) \;
    }
}

\textbf{function Select\_$Z$}($server$, $Z_c^{t_i}$):\\ 
\Indp  
\For{$Z_{sq}^{t_i} \in server$}{
    $similarities \gets \textbf{cosine\_similarity}(Z_{sq}^{t_i}, Z_c^{t_i})$ \; 
    \If{$\sum(similarities > \alpha) \geq \frac{\text{len}(similarities)}{2}$}{
        $selected\_z$.append($Z_{sq}^{t_i}$) \; 
    }
    }

\Return $selected\_z$ \; 
\Indm 

\end{algorithm}
Furthermore, to define the MKVs and the re-editing conditions, we summarize our framework FedEdit in Algorithm 1, which consists of two main steps: 

\textbf{(i) Editing.} Between the time slots in $T$, each client executes the Edit algorithm parallelly and independently (Step (1)). Here we still take MEMIT as an example i.e., Algorithm \ref{alg:MEMIT} in Appendix \ref{sec:appendixA}. In this algorithm process, $z_{ci}$, $k_{ci}^{l_0}$ are related to the data records $\mathcal{E}_c$ , and we define the mediator knowledge vectors (MKVs) of client $c$ at time $t$ as $Z_c^t$:
\begin{equation}
Z_c^t = \left\{(z_{ci}, k_{ci}^{L_0}) \right\}
\end{equation}
where $(z_{ci}, k_{ci}^{L_0})$ are all generated by client $c$ during the time interval from $t-1$ to $t$, $(s_{ci}, r_{ci}, o_{ci}) \in \mathcal{E}_c$, and the keys $k_{ci}^{L_0}$ at the $l$-th layer are defined as follows \cite{r2}:
\begin{equation}
k_{ci}^{L_0} = \frac{1}{P} \sum_{j=1}^{P} k(x_{cj} + s_{ci}),
\end{equation}
where $k(x) = \sigma(W_{I}^l \gamma (h_{ci}^{L_0-1}(x)))$. Once a client has finished editing, it uploads the obtained $Z_c$ to the server (Step (2)).

\textbf{(ii) Re-editing.} Once the time reaches any time $t_i \in T (i = 1, ... , m, m$ is the total number of time slots$)$, where $m$ is the server $s$ distributes the previously stored $Z_s^{t_i}$ between $t_{i-1}$ to $t_i$ to each client. Each client selects $Z_c^{t_i}$ from the $Z_s^{t_i}$ that are beneficial to it, i.e., positively correlated with its own data $\mathcal{E}_c$ indirectly, through the \textbf{``re-edit'' condition}:
\begin{equation}
\sum(\text{\textit{similarities}} > \alpha) \geq \frac{\text{\textit{len}(\textit{similarities})}}{2}
\end{equation}
where \textit{similarities} is the cosine similarity between the $q$-th traversed $Z_{sq}^{t_i}$ in the server and $Z_c^{t_i}$, i.e., line 10 of Algorithm \ref{alg:FedEdit}. $\alpha$ means similarity threshold, which is a hyperparameter. $\sum(\text{\textit{similarities}} > \alpha)$ is the number of $Z_{sq}^{t_i}$ in $Z_s^{t_i}$ that satisfy the similarity threshold requirement. $\textit{len}(\textit{similarities})$ is the number of all $(z_{ci}, h_{ci}^L, k_{ci}^L)$ in the client $c$ as of the current time slot $t$. In summary, iterates through each $Z_{sq}^{t_i}$ in the server, calculates the cosine similarity between the $Z_{sq}^{t_i}$ and the $Z_c^{t_i}$ of client $c$, and if more than half of the $(z_{ci}, h_{ci}^L, k_{ci}^L)$'s in client $c$ are greater than the similarity threshold $\alpha$, then the $Z_{sq}^{t_i}$ is said to satisfy the current client's ``re-edit'' condition. Then the $Z_{sq}^{t_i}$ will be selected by client $c$ (Step (3)).

When the screening process is finished, each client performs Algorithm \ref{alg:MEMIT} again on the basis of the model ${edited\_model}_c$ that has been edited earlier (Step (4)). The process is repeated until $t_i=t_m$.

\section{Experiments}
\begin{table*}[htbp]
\centering
\scriptsize
\begin{tabular}{ccccccc}
\toprule
\textbf{Editor} & \textbf{Score} & \textbf{Efficacy} & \textbf{Generalization} & \textbf{Specificity} & \textbf{Fluency} & \textbf{Consistency} \\
\midrule
GPT-J (6B) & 22.4 & 15.2 (0.7) & 17.7 (0.6) & 83.5 (0.5) & 622.4 (0.3) & 29.4 (0.2) \\
\midrule
\colorbox{nonFed}{FT-W} & 67.6 & 99.4 (0.1) & 77.0 (0.7) & 46.9 (0.6) & 293.9 (2.4) & 15.9 (0.3) \\
\colorbox{nonFed}{MEND} & 23.1 & 15.7 (0.7) & 18.5 (0.7) & 83.0 (0.5) & 618.4 (0.3) & 31.1 (0.2) \\
\colorbox{nonFed}{ROME} & 50.3 & 50.2 (1.0) & 50.4 (0.8) & 50.2 (0.6) & 589.6 (0.5) & 3.3 (0.0) \\
\midrule
\colorbox{nonFed}{MEMIT} & 85.8 & 98.9 (0.2) & 92.8 (0.4) & 73.7 (0.5) & 619.9 (0.3) & 40.1 (0.2) \\
\colorbox{Fed}{MEMITAvg} & 62.9 [73.3\%] & 60.4 [61.1\%] & 55.5 [59.8\%] & \textbf{76.5 [103.8\%]} & \textbf{619.8 [99.9\%]} & 35.2 [87.8\%] \\
\colorbox{Fed}{FedMEMIT} & \textbf{85.1 [99.2\%]} & \textbf{97.0 [98.1\%]} & 86.5 [93.2\%] & \textbf{75.4 [102.3\%]} & \textbf{614.9 [99.2\%]} & 37.8 [94.3\%] \\
\midrule
\colorbox{nonFed}{PMET} & 86.2 & 99.5 (0.1) & 88.6 (0.4) & 71.4 (0.5) & 620.0 (0.3) & 40.6 (0.2) \\
\colorbox{Fed}{PMETAvg} & 35.9 [41.6\%] & 28.4 [28.5\%] & 27.8 [31.4\%] & \textbf{80.8 [113.2\%]} & \textbf{623.2 [100.5\%]} & 31.5 [77.6\%] \\
\colorbox{Fed}{FedPMET} & \textbf{83.6 [97.0\%]} & \textbf{98.9 [99.4\%]} & \textbf{93.2 [105.2\%]} & 67.2 [94.1\%] & \textbf{619.5 [99.9\%]} & \textbf{39.6 [97.5\%]} \\
\midrule
\midrule
GPT-NeoX (20B) & 23.7 & 16.8 (1.9) & 18.3 (1.7) & 81.6 (1.3) & 620.4 (0.6) & 29.3 (0.5) \\
\midrule
\colorbox{nonFed}{MEMIT} & 82.0 & 97.2 (0.8) & 82.2 (1.6) & 70.8 (1.4) & 606.4 (1.0) & 36.9 (0.6) \\
\colorbox{Fed}{MEMITAvg} & 37.7 [46.0\%] & 30.7 [31.6\%] & 29.4 [35.8\%] & \textbf{77.3 [109.2\%]} & \textbf{618.0 [101.9\%]} & 30.9 [83.7\%] \\
\colorbox{Fed}{FedMEMIT} & \textbf{80.8 [98.5\%]} & \textbf{96.9 [99.7\%]} & \textbf{89.6 [109.0\%]} & 64.1 [90.5\%] & \textbf{598.5 [98.7\%]} & \textbf{40.8 [110.6\%]} \\
\midrule
\colorbox{nonFed}{PMET} & 84.3 & 98.4 (0.2) & 89.4 (0.5) & 70.3 (0.5) & 598.1 (0.6) & 38.9 (0.2) \\
\colorbox{Fed}{PMETAvg} & 36.2 [43.0\%] & 29.0 [29.5\%] & 28.0 [31.1\%] & \textbf{79.7 [113.4\%]} & \textbf{618.4 [103.4\%]} & 30.8 [79.2\%] \\
\colorbox{Fed}{FedPMET} & \textbf{84.3 [100\%]} & \textbf{95.6 [97.2\%]} & \textbf{91.3 [102.1\%]} & \textbf{70.7 [100.6\%]} & \textbf{579.5 [96.9\%]} & 34.1 [87.7\%] \\
\bottomrule
\end{tabular}
\caption{
10,000 counterfact edits on GPT-J (6B) and GPT-NeoX (20B) in \colorbox{Fed}{federated} and \colorbox{nonFed}{centralized} scenarios. Parentheses indicate the 95\% confidence interval, while brackets show federated scenario metrics as a percentage of the centralized scenario, with values exceeding 95\% \textbf{bolded}.
}
\label{tab:mainTable}
\end{table*}
\subsection{Experimental Setup}
\textbf{Datasets.} We conducted counterfactual update experiments on two datasets: Zero-Shot Relation Extraction (zsRE) \cite{r4} and C{\small{OUNTER}}F{\small{ACT}} \cite{r5}. The zsRE dataset contains 10,000 real-world facts \cite{r2}, while C{\small{OUNTER}}F{\small{ACT}} includes 21,919 factual statements \cite{r5}. To simulate FLEKE, we reorganized the datasets using different clustering methods. For zsRE, we clustered data based on the "src" value, which represents the subject (e.g., "What university did Watts Humphrey attend?" with the subject "Watts Humphrey"). We applied spectral clustering after transforming the text into word vectors to assign data to different clients. For C{\small{OUNTER}}F{\small{ACT}}, we grouped data with the same "relation id" into one client, and randomly assigned about 1/10 of the data from other clients to each client. \\
\textbf{Baselines.}
We select six knowledge editing methods as baselines:
(1) \textbf{FT-W} is a simple fine-tuning approach that applies weight decay to prevent forgetfulness.
(2) \textbf{MEND} \cite{r6} transforms the fine-tuning gradient of an updated fact by decomposing the weight matrix into rank-1 form using a pre-trained hyper-network.
(3) \textbf{ROME} \cite{r5} locates factual retrievals within a specific set of MLP modules and updates knowledge by directly writing new key-value pairs into the MLP module.
(4) \textbf{MEMIT} \cite{r2} extends ROME to insert multiple memories by modifying the MLP weights of several critical layers.
(5) \textbf{PMET} \cite{r7} uupdates FFN weights by optimizing the hidden states of both MHSA and FFN, using only the FFN hidden states for weight updates.
(6) \textbf{EditAvg} is a variant of FedAvg for solving the FLEKE task, where any LEKE method can replace "Edit." Please refer to Appendix~\ref{EditAvg} for the detail settings. \\
\textbf{Metrics.} Following \citeauthor{r5} (\citeyear{r5}), we use GPT-J (6B) \cite{r20} and GPT-NeoX (20B) \cite{r1} as the backbone for FLEKE. 
Following prior work \cite{r2}, we evaluate models using the following metrics: (1) Efficacy, measuring editing success; (2) Paraphrase, assessing success on rephrasings of the original statement; (3) Specificity, ensuring unrelated facts remain unchanged; and (4) Score, the harmonic mean of these three metrics, balancing reliability (efficacy and paraphrase) and specificity. Additionally, in C{\small{OUNTER}}F{\small{ACT}} experiments, we include (5) Fluency, evaluating degradation due to repetition, and (6) Consistency, measuring semantic coherence in generated text. All results are weighted averages across clients. \\
\textbf{Hyper-parameters.} 
We set the number of clients to 8, with a total of approximately 10,000 edits, and define T to consist of 10 time slots. Covariance statistics are collected on GPT-J using 100,000 samples from Wikitext, and on GPT-NeoX using 50,000 samples from Wikitext. Please refer to Appendix~\ref{hyperparameter} for more details. 

\subsection{Results of C{\small{OUNTER}}F{\small{ACT}}}
Table \ref{tab:mainTable} presents the results of all methods on 10K counterfactual edits. FedMEMIT and FedPMET achieve 99.2\% and 97\% of the performance of centralized methods, respectively. In contrast, applying the FedAvg algorithm to MEMIT and PMET results in only 73.3\% and 41.6\%, respectively. This demonstrates that our method performs well in FLEKE. It also highlights that simply combining federated learning algorithms like the classical FedAvg with knowledge editing methods does not yield effective results. In the trade-off between editing reliability and specificity, FedMEMIT and FedPMET, like MEMIT and PMET, prioritize reliability. On the other hand, MEND, MEMITAvg, and PMETAvg focus more on specificity. Moreover, FedMEMIT and FedPMET outperform non-federated methods in terms of specificity and generalization, respectively. However, in terms of specificity, they fall behind the meta-learning-based method MEND.

Next, we applied the FedEdit framework to perform 10K edits on GPTNeoX (20B) using the C{\small{OUNTER}}F{\small{ACT}} dataset. The results are shown in the lower part of Table \ref{tab:mainTable}. We find: FedMEMIT and FedPMET significantly outperform MEMITAvg and PMETAvg, consistently favoring reliability and consistency. Additionally, both FedMEMIT and FedPMET surpass their respective non-federated methods in generalization. This may be due to our proposed “re-edit” condition, which selects data with similar types for re-editing, thereby enhancing reliability. We further explore this in the following ablation experiments.

\begin{table}[t]
\centering
\resizebox{\linewidth}{!}{
\begin{tabular}{@{}lcccc@{}}
\toprule
\textbf{Editor} & \textbf{Score} & \textbf{Efficacy} & \textbf{Generalization} & \textbf{Specificity} \\ \midrule
GPT-J           & 26.4   & 26.4 ($\pm$0.6)   & 25.8 ($\pm$0.5)         & 27.0 ($\pm$0.5)      \\
\colorbox{nonFed}{FT-W}            & 42.1   & 69.6 ($\pm$0.6)   & 64.8 ($\pm$0.6)         & 24.1 ($\pm$0.5)      \\
\colorbox{nonFed}{MEND}            & 20.0   & 19.4 ($\pm$0.5)   & 18.6 ($\pm$0.5)         & 22.4 ($\pm$0.5)      \\
\colorbox{nonFed}{ROME}            & 2.6   & 21.0 ($\pm$0.7)   & 19.6 ($\pm$0.7)         & 0.9 ($\pm$0.1)       \\
\midrule
\colorbox{nonFed}{MEMIT}           & 50.7   & 96.7 ($\pm$0.3)   & 89.7 ($\pm$0.5)         & 26.6 ($\pm$0.5)      \\
\colorbox{Fed}{MEMITAvg}        & 41.6 [82.1\%]      & 55.7 [57.6\%]      & 53.7 [59.9\%]           & 28.1 [105.6\%]      \\
\colorbox{Fed}{FedMEMIT}        & \textbf{50.5 [99.6\%]}     & \textbf{92.9 [96.1\%]}      & \textbf{87.3 [97.3\%]}           & \textbf{26.9 [101.1\%]}      \\
\midrule
\colorbox{nonFed}{PMET}            & 51.0   & 96.9 ($\pm$0.3)   & 90.6 ($\pm$0.2)         & 26.7 ($\pm$0.2)      \\
\colorbox{Fed}{PMETAvg}         & 41.5 [81.4\%]     & 55.5 [57.3\%]      & 53.3 [58.8\%]           & \textbf{28.2 [105.6\%]}      \\
\colorbox{Fed}{FedPMET}         & 42.5 [82.4\%]     & 66.5 [68.6\%]      & 61.8 [68.2\%]                   & \textbf{25.4 [95.1\%]}               \\
\bottomrule
\end{tabular}}
\caption{10,000 zsRE Edits on GPT-J (6B).}
\label{tab:zsRE}
\vspace{-0.1in}
\end{table}

\subsection{Results of ZsRE}
The zsRE dataset tests the ability to add correct information. The results of editing 10K knowledge on the zsRE dataset are shown in Table \ref{tab:zsRE}. These results demonstrate that our method performs very close to the original method in the federated scenario, both in efficacy and generalization metrics, and even slightly outperforms it in terms of specificity.
Specificity refers to the model’s argmax accuracy on a randomly sampled, unrelated fact that should not have changed \cite{r2}. EditAvg (MEMITAvg and PMETAvg), averages the $\Delta_c^t$ of each client before inserting it into the server's model, making it naturally stable in the case of random sampling. Additionally, compared to the original method, our approach includes an extra re-editing step. This step allows for editing additional vectors that are more suitable for the current client, improving performance.

\subsection{Ablation Study}

\begin{table}[t]
\centering
\resizebox{\linewidth}{!}{
\begin{tabular}{lcccccc}
\toprule
\textbf{Edits} & \textbf{Editor} & \textbf{Score} & \textbf{Efficacy} & \textbf{Generalization} & \textbf{Specificity} \\
\midrule
 & GPT-J & 26.4 & 25.8 & 27.0 & 26.4 \\
\midrule
1K & \colorbox{Fed}{FedMEMIT}  & 57.0 & 99.7 & 97.1 & 31 \\
 & \colorbox{Fed}{w/o $Z^t_c$} & 54.1 (↓2.9) & 99.6 (↓0.1) & 97.2 (↑0.1) & 28.5 (↓2.5) \\
 & \colorbox{Fed}{FedPMET} & 54.9 & 98.0 & 94.0 & 29.6 \\
 & \colorbox{Fed}{w/o $Z^t_c$} & 54.3 (↓0.6) & 97.4 (↓0.6) & 93.6 (↓0.4) & 29.2 (↓0.4) \\
\midrule
5K & \colorbox{Fed}{FedMEMIT} & 55.1 & 98.2 & 94.0 & 29.8 \\
 & \colorbox{Fed}{w/o $Z^t_c$} & 53.2 (↓1.9) & 96.1 (↓2.1) & 91.6 (↓2.4) & 28.5 (↓1.3) \\
 & \colorbox{Fed}{FedPMET} & 52.8 & 93.6 & 88.4 & 28.7 \\
 & \colorbox{Fed}{w/o $Z^t_c$} & 51.9 (↓0.8) & 91.6 (↓2.0) & 85.6 (↓2.8) & 28.4 (↓0.3) \\
\midrule
10K & \colorbox{Fed}{FedMEMIT} & 50.5 & 92.9 & 87.3 & 26.9 \\
 & \colorbox{Fed}{w/o $Z^t_c$} & 49.6 (↓0.9) & 87.7 (↓5.2) & 83.0 (↓4.3) & 27.0 (↑0.1) \\
 & \colorbox{Fed}{FedPMET} & 42.5 & 66.5 & 61.8 & 25.4 \\
 & \colorbox{Fed}{w/o $Z^t_c$} & 40.8 (↓1.7) & 65.5 (↓1.0) & 60.5 (↓1.3) & 24.0 (↓1.4) \\
\bottomrule
\end{tabular}}
\caption{The ablation experiments, where w/o $Z_c^t$ means that the re-editing condition control is removed.} 
\label{tab:ablation}
\vspace{-0.1in}
\end{table}

The ablation study in Table \ref{tab:ablation} examines the impact of removing the re-editing condition control from the FedMEMIT and FedPMET methods on their performance across different editing scales (1K, 5K, and 10K edits). The results show that: (1) The re-editing condition is crucial for the FLEKE task. Removing it causes a decline in the score for all FedEdit-related experiments, indicating that the model updates facts unrelated to its own data. This negatively impacts the model’s ability to edit its own knowledge accurately. (2) The more similar the knowledge edited on a single client, the better the model's reliability. A clear trend emerges: reliability (efficacy and paraphrasing) decreases more, while specificity decreases less. This suggests that as the number of client edits increases, the re-editing condition improves reliability.

\subsection{Robustness Study}
\begin{figure}[t]
 \centering
\subfigure{ 
 \includegraphics[width=0.22\textwidth]{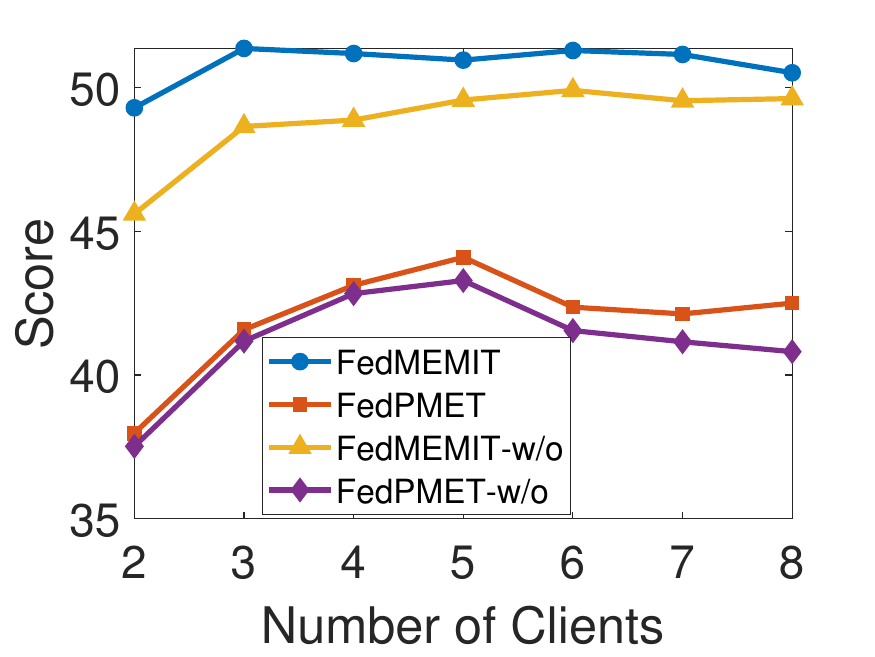}
 }
 \subfigure{
 \includegraphics[width=0.22\textwidth]{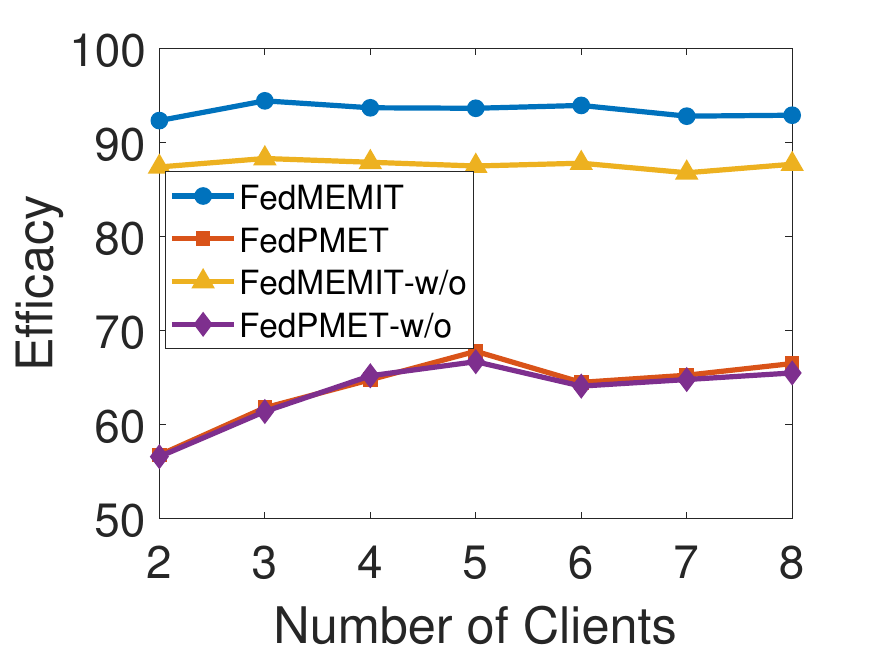}
 }
 \subfigure{
 \includegraphics[width=0.22\textwidth]{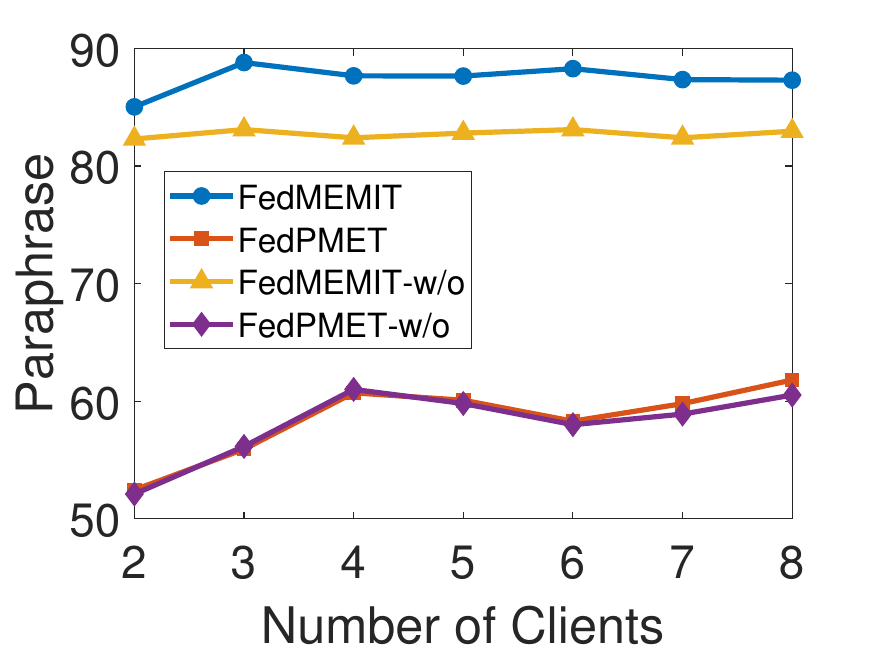}
 }
 \subfigure{
 \includegraphics[width=0.22\textwidth]{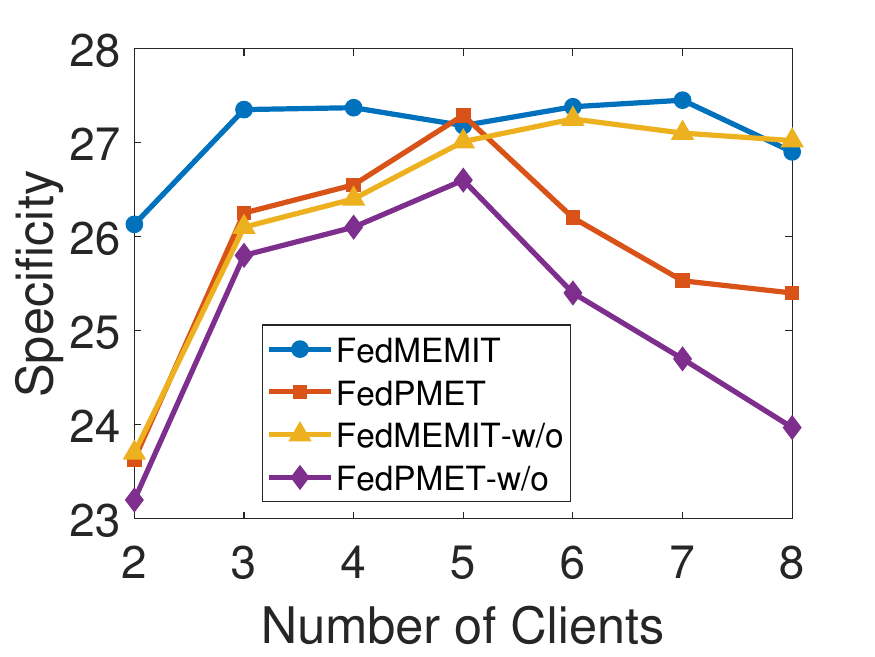}
 }
 \caption{The editing performance of FedEdit and baselines with the number of clients. The suffix "w/o" indicates the Ablation experimental group. }
 \label{fig:impactOfClient}
 \vspace{-0.1in}
\end{figure}
We conducted a robustness study on the proposed framework using the zsRE dataset. Specifically, we assigned 10,000 facts to {2, 3, 4, 5, 6, 7, 8} clients for editing. As shown in Fig.~\ref{fig:impactOfClient}, the experimental charts show that as the number of clients increases, FedMEMIT consistently performs well and remains stable across all metrics. In contrast, while FedPMET's performance improves initially, it declines as the number of clients grows, likely due to the effects of multiple edits. Notably, the Specificity indicator fluctuates with the number of clients, which may be influenced by the number of single edits.

\subsection{Impact of the Number of Time Slots}
\begin{figure}[t]
 \centering
\subfigure{
 \includegraphics[width=0.22\textwidth]{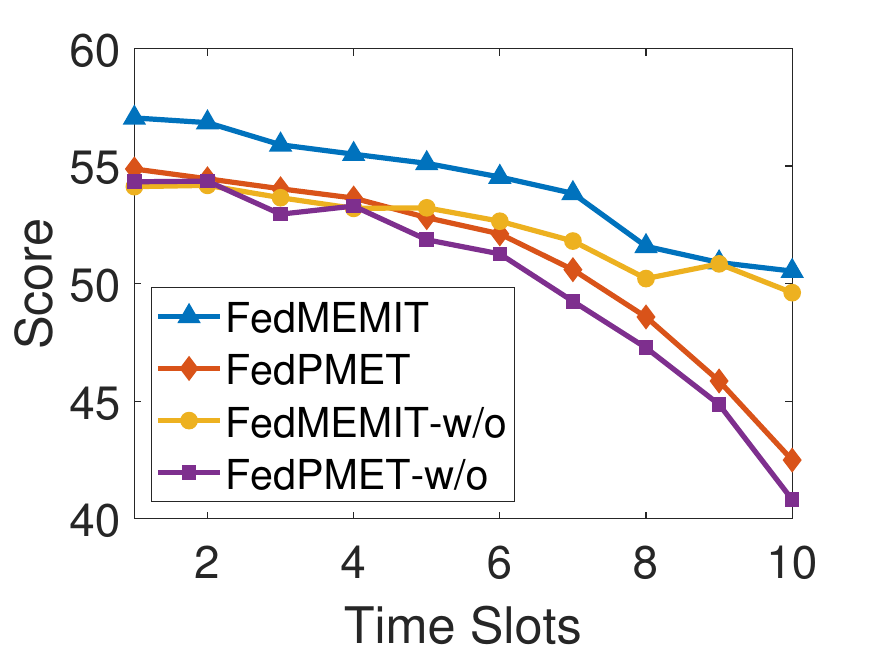}
 }
 \subfigure{
 \includegraphics[width=0.22\textwidth]{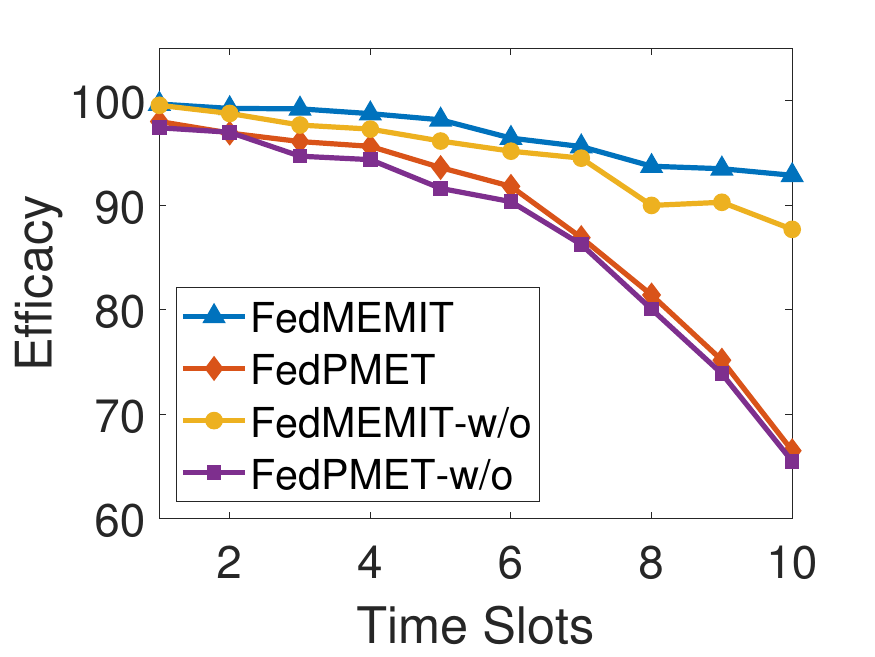}
 }
 \subfigure{
 \includegraphics[width=0.22\textwidth]{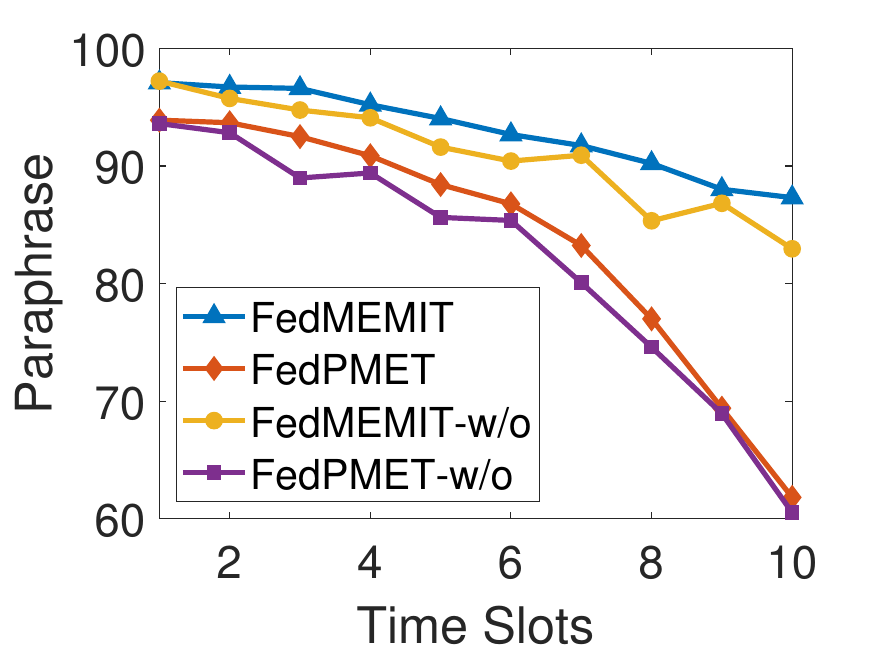}
 }
 \subfigure{
 \includegraphics[width=0.22\textwidth]{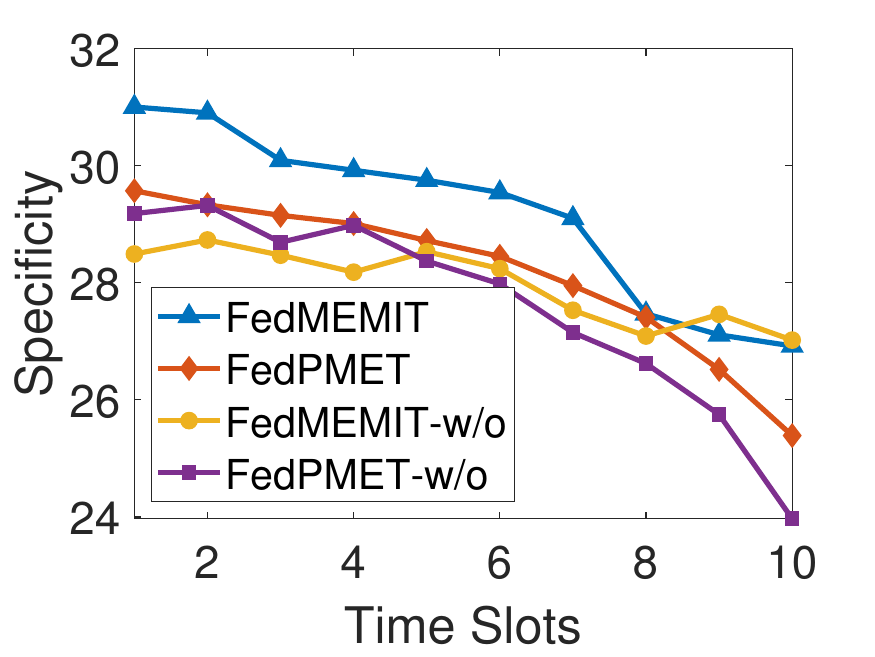}
 }
 \caption{The editing performance of FedEdit and baselines with the number of time slots.}
 \label{fig:impactOfTime}
 \vspace{-0.1in}
\end{figure}
Fig.~\ref{fig:impactOfTime} illustrates the superior performance of FedMEMIT compared to FedPMET and the ablation variants, FedMEMIT-w/o and FedPMET-w/o. While both FedMEMIT and FedPMET show a decline in performance as the number of time slots increases, FedMEMIT consistently outperforms FedPMET across all metrics. FedMEMIT achieves higher scores, maintains better efficiency, and demonstrates more stable paraphrasing quality and specificity, especially in long-term editing tasks. The gradual decline in FedMEMIT indicates its better ability to preserve edit quality over time, compared to the more significant performance drops seen in FedPMET. This highlights FedMEMIT’s robustness, showing advantages in efficiency, editing reliability, and specificity preservation. Although FedMEMIT-w/o shows improvements in stability, FedMEMIT remains the most effective for achieving high-quality, sustainable editing performance in federated scenarios.

\subsection{Case Study}
Table \ref{tab:case} demonstrates that FedMEMIT correctly generates text in both cases, in contrast to MEMITAvg. This highlights the limitations of selecting $\Delta_c$ as the mediator vector and validates the appropriateness of choosing $Z_c$ as the mediator vector. Moreover, the table illustrates a scenario where two highly similar cases are edited on two different clients. Specifically, case 15874 on client 1 is first edited, and the resulting $Z_c$ vector is uploaded to the server. Client 2 then retrieves this vector from the server and edits it. As a result, when a similar case (case id: 15911) is edited, the text is generated correctly. However, if the vector has not been edited, client 2 generates incorrect text. This further demonstrates the effectiveness of FedEdit.

\begin{table}[t]
\centering
\resizebox{\linewidth}{!}{
\begin{tabular}{@{}ll@{}}
\toprule
\textbf{Client 1:(case id: 15874)} \\
\textbf{Original Knowledge:} Josef Albers, who has a citizenship from Germany \\
\textbf{Edited Knowledge:} Josef Albers, who has a citizenship from Canada \\
\midrule
\textbf{MEMITAvg:} \underline{\textit{Josef Albers, who has a citizenship from}} \textcolor{red}{?0.0}. \\
\textbf{FedMEMIT:} \underline{\textit{Josef Albers, who has a citizenship from}} \textcolor{green}{Canada}. \\
\midrule
\midrule
\textbf{Client 2:(case id: 15911)} \\
\textbf{Original Knowledge:} Ulrica Arfvidsson, who holds a citizenship from Sweden \\
\textbf{Edited Knowledge:} Ulrica Arfvidsson, who holds a citizenship from Kenya \\
\midrule
\textbf{MEMITAvg:} \underline{\textit{Ulrica Arfvidsson, who holds a citizenship from}} \textcolor{red}{Sweden}. \\
\textbf{FedMEMIT (Before 15874):}\underline{\textit{Ulrica Arfvidsson, who holds a citizenship from}} \textcolor{red}{Italy}. \\
\textbf{FedMEMIT (After 15874):} \underline{\textit{Ulrica Arfvidsson, who holds a citizenship from}} \textcolor{green}{Kenya}. \\
\bottomrule
\end{tabular}}
\caption{Results for two cases in C{\small{OUNTER}}F{\small{ACT}} from two clients based on GPT-J (6B). Prompt $+$ subject are underlined and italicized. Words highlighted in \textcolor{green}{green} signify keywords that reflect correct behavior. Those in \textcolor{red}{red} denote keywords associated with incorrect behavior.}
\label{tab:case}
\vspace{-0.1in}
\end{table}

\section{Conclusion}
We introduce FLEKE, a novel task that enables collaborative knowledge editing across multiple clients while ensuring privacy and reducing computational costs. To achieve this, we propose FedEdit, a two-stage framework comprising editing and re-editing. In the editing stage, clients locally perform knowledge editing and upload MKVs to a central server. In the re-editing stage, clients retrieve relevant MKVs via cosine similarity for further refinement. Experimental results demonstrate that FedEdit outperforms strong baselines in FLEKE, paving the way for more effective knowledge editing in federated settings and inspiring future research in this direction.

\section*{Limitation}
We acknowledge the following limitations in our work: (1) The FLEKE task may face challenges due to non-IID data across clients. The heterogeneous data distributions can cause instability in the model, particularly when personalization is required for different tasks. While we have addressed this issue through the FedEdit framework, which uses clustering for selecting MKVs and re-editing conditions to improve the knowledge editing process, it remains a challenge in environments with diverse data. (2) Our work focuses on a simulated federated learning scenario, and thus does not account for certain external factors, such as environmental changes or system anomalies, that may impact the performance of the deployment in real-world settings. We plan to conduct additional experiments to further explore these challenges.

\section*{Ethics Consideration}
In the development and application of federated learning systems, we prioritize ethical sourcing and privacy protection. Our proposed FLEKE task ensures that the research complies with data privacy regulations, and all datasets used in this study (zsRE and C{\small{OUNTER}}F{\small{ACT}}) are open-source and publicly available. These datasets do not contain any personally identifiable information or sensitive data. To mitigate privacy risks, our proposed FedEdit framework ensures that only mediator knowledge vectors (MKVs) are uploaded to the server, rather than raw data. This design ensures that sensitive data is never directly shared, and knowledge editing is performed in a manner that prevents leakage of private information.

Additionally, while federated learning frameworks enable collaboration among different organizations, we acknowledge the importance of safeguarding intellectual property and ensuring fairness in model training. Our work is designed to facilitate efficient knowledge editing while preventing misuse or unintended consequences. As such, we have implemented careful oversight measures to ensure that the server-based aggregation of MKVs does not inadvertently expose confidential information.

Furthermore, all experiments were conducted with transparency and respect for the principles of fairness and data protection. We do not authorize the use of the datasets for any commercial purposes, and our results are strictly intended for academic and research purposes. Our study demonstrates the potential of federated learning to enhance the efficiency and privacy of knowledge editing tasks, while adhering to ethical standards of data use and model deployment.

\section*{Acknowledgements}
This work is supported in part by Science and Technology Innovation Key R\&D Program of Chongqing (Grant No. CSTB2023TIAD-STX0035), National NSFC (Grants No. 62372072, 62102053, and 52272388), Chongqing Talent Program Contract System Project (Grant No. cstc2024ycjh-bgzxm0042), Haihe Lab of ITAI (Grant No. 22HHXCJC00002), the Natural Science Foundation of Chongqing, China (Grant No. CSTB2022NSCQ-MSX1104), Key Laboratory of Big Data Intelligent Computing, Chongqing University of Posts and Telecommunications (Grant No. BDIC-2023-B-003), Regional Innovation Cooperation Project of Sichuan Province (Grants No. 2023YFQ0028 and 2024YFHZ0097), Regional Science and Technology Innovation Cooperation Project of Chengdu City (Grant No. 2023-YF11-00023-HZ), and China Postdoctoral Science Foundation Funded Project (2024M763867).
\bibliography{custom}

\appendix
\section{Federated Learning}
\label{sec:appendixB}
\textbf{Training Objective.} Federated learning aims to optimize the following objective function:
\begin{equation} \label{Eq:fed-loss}
    \begin{aligned}
        \min_{w} \mathcal{F}(w) &\triangleq \sum_{i=1}^{N} p_i \mathcal{L}_i(w) \\
        \text{where} \quad \mathcal{L}_i(w) &= \mathbb{E}_{a \sim \mathcal{D}_i}[f_i(w, a)].
    \end{aligned}
\end{equation}
 
In Eqn.(\ref{Eq:fed-loss}), $\mathcal{L}_i(w)$ denotes the local training objective function of the client $i$ and $N$ denotes the number of clients. $w \in \mathbb{R}^d$ represents the parameters of the global model. $a$ denotes each batch of data. The local training loss function $f_i(w, a)$ is  often the same across all the clients, while $\mathcal{D}_i$ denotes the distribution of the local client data, which is often different across the clients, capturing the heterogeneity. $p_i$ is defined as the training size proportion in Eqn. (4),  where $|\mathcal{D}_i|$ is the training size  of client $i$.
\begin{equation}
    p_i = \frac{|\mathcal{D}_i|}{\sum_{i=1}^{N} |\mathcal{D}_i|}
\end{equation}
\textbf{Training Procedure.} Federated learning is an iterative process shown in Figure 2. The server initializes the global model, followed by multiple communication rounds between the server and clients. In each \textit{communication round}, there are four steps between the server and clients. 1) In round $t$, the server sends the global model $w^t$ to all the clients. 2) After clients receive the global model $w^t$ as the initialization of the local model, they start to train it using their own data for multiple epochs and obtain the local model changes $\Delta w_i^t$ during the local  training stage. 3) The clients send their local model changes to the server. 4) The server aggregates the local model changes $\Delta w_i^t$ collected from different  clients as Eqn. (3) shows, and then uses the $t$-$th$ round’s global model $w^t$ and the aggregated local model changes $\Delta w_i^t$ to update the global model. As Eqn. (4) shows, $w^{t+1}$ is the global model after the update. Here, $\mathrm{n}$ denotes the server learning rate. The server will send the updated model $w^{t+1}$ to the clients, then the $(t+1)$-$th$ round starts.

The above procedure will repeat until the algorithm converges.
\begin{align}
    \Delta w^t &= \sum_{i=1}^{N} p_i \Delta w_i^t \label{eq:delta_w} \\ 
    w^{t+1} &= w^t - \eta \Delta w^t \label{eq:update_w}
\end{align}
\textbf{FedAvg.} Federated Averaging (FedAvg) \cite{r3} uses stochastic gradient descent (SGD) as the local training optimizer to optimize the training procedure and uses the same learning rate and the same number of local training epochs for all the clients.

\section{Details of EditAvg} \label{EditAvg}
In this approach, $\Delta_c^t$ is selected as the MKV for client $c$ at time $t \in T$, and it is transferred to the server after execution of Algorithm \ref{alg:MEMIT} (equation (\ref{Eq:Delta})). The server then aggregates all $\Delta_c^t$ using the formula $\Delta_c^t = \sum_{c=1}^{N} p_c^t \Delta_c^t$, where $p_c^t$ represents the proportion of edits made by client $c$ from $t-1$ to $t$. \\
\section{MEMIT}
\label{sec:appendixA}

\begin{algorithm}
\caption{MEMIT} \label{alg:MEMIT}
\KwIn{Requested edits $\mathcal{E} = \{(s_i, r_i, o_i)\}$, generator $G$, layers to edit $S$, covariances $C^l$}
\KwOut{Modified generator containing edits from $\mathcal{E}$}

\For{$s_i, r_i, o_i \in \mathcal{E}$}{
    $\delta_i \gets $$argmin_{\delta_i} \frac{1}{P} \sum_{j=1}^P -\log \mathbb{P}_{G(h_i^L += \delta_i)}$
    $\left[o_i \mid x_j \oplus p(s_i, r_i)\right]$\;
    $z_i \gets h_i^L + \delta_i$\;
}
\For{$l \in \mathcal{R}$}{
    $h_i^l \gets h_i^{l-1} + a_i^l + m_i^l$\;
    \For{$s_i, r_i, o_i \in \mathcal{E}$}{
        $k_i^l \gets \frac{1}{P} \sum_{j=1}^P k(x_j + s_i)$\;
        $r_i^l \gets \frac{z_i - h_i^L}{L - l + 1}$\;
    }
    $K^l \gets \{k_i^l, \dots, k_i^l\}$\;
    $R^l \gets \{r_i^l, \dots, r_i^l\}$\;
    $\Delta^l \gets R^l K^{l^T} (C^l + K^l K^{l^T})^{-1}$\;
    $W^l \gets W^l + \Delta^l$ 
}
\end{algorithm}
In Algorithm \ref{alg:MEMIT}, with the exception of the symbol $r_i^l$ in line 8, the symbol definitions in the rest of the formulas are consistent with those defined in section \ref{method: method}. $r_i^l \gets \frac{z_i - h_i^L}{L - l + 1}$ is the residual, which is spread over layers $\mathcal{R}$.

\section{Detailed Hyper-parameters} \label{hyperparameter}
We set the number of clients to 8, with a total of approximately 10,000 edits, and define T to consist of 10 time slots. Covariance statistics are collected on GPT-J using 100,000 samples from Wikitext, and on GPT-NeoX using 50,000 samples from Wikitext. $\mathcal{R}=\{3,4,5,6,7,8\}$ for GPT-J and $\mathcal{R}=\{6,7,8,9,10\}$ for GPT-NeoX. Further implementation details about LKE are the same as \citeauthor{r2} (\citeyear{r2}) and \citeauthor{r7} (\citeyear{r7}). For the computing resources, we utilize 8 NVIDIA A800 80GB GPUs.
\end{document}